\documentclass[sigconf,authorversion]{acmart}

\usepackage{algorithm}
\usepackage{multirow}
\usepackage{appendix}
\usepackage{algorithmic}

\usepackage{enumitem} 
\usepackage{subcaption}

\AtBeginDocument{%
  }

\setcopyright{acmcopyright}
\copyrightyear{2024}
\acmYear{2024}
\acmDOI{XXXXXXX.XXXXXXX}

\acmConference[ETRA'24]{ETRA 2024}{June 04--07,
  2024}{Glasgow, United Kingdom}

\title{A Transformer-Based Model for the Prediction of Human Gaze Behavior on Videos}

\copyrightyear{2024}
\acmYear{2024}
\setcopyright{acmlicensed}\acmConference[ETRA '24]{2024 Symposium on Eye Tracking Research and Applications}{June 4--7, 2024}{Glasgow, United Kingdom}
\acmBooktitle{2024 Symposium on Eye Tracking Research and Applications (ETRA '24), June 4--7, 2024, Glasgow, United Kingdom}
\acmDOI{10.1145/3649902.3653439}
\acmISBN{979-8-4007-0607-3/24/06}

\begin{document}

\author{Suleyman Ozdel}
\authornote{Both authors contributed equally to the paper.}
\affiliation{%
  \institution{Technical University of Munich}
  \city{Munich}
  \country{Germany}
}
\email{ozdelsuleyman@tum.de}
\orcid{0000-0002-3390-6154}

\author{Yao Rong}
\authornotemark[1]
\affiliation{%
  \institution{Technical University of Munich}
  \city{Munich}
  \country{Germany}
}
\email{yao.rong@tum.de}
\orcid{0000-0002-6031-3741}

\author{Berat Mert Albaba}
\affiliation{%
  \institution{ETH}
  \city{Zurich}
  \country{Switzerland}
}
\email{mert.albaba@inf.ethz.ch}
\orcid{0000-0002-3406-8412}

\author{Yen-Ling Kuo}
\affiliation{%
  \institution{University of Virginia}
  \city{Charlottesville}
  \country{United States}
}
\email{ylkuo@virginia.edu}
\orcid{0000-0002-6433-6713}

\author{Xi Wang}
\affiliation{%
  \institution{ETH}
  \city{Zurich}
  \country{Switzerland}
}
\email{xi.wang@inf.ethz.ch}
\orcid{0000-0001-5442-1116}

\author{Enkelejda Kasneci}
\affiliation{%
  \institution{Technical University of Munich}
  \city{Munich}
  \country{Germany}}
\email{enkelejda.kasneci@tum.de}
\orcid{0000-0003-3146-4484}

\begin{abstract}
Eye-tracking applications that utilize the human gaze in video understanding tasks have become increasingly important. To effectively automate the process of video analysis based on eye-tracking data, it is important to accurately replicate human gaze behavior. However, this task presents significant challenges due to the inherent complexity and ambiguity of human gaze patterns. In this work, we introduce a novel method for simulating human gaze behavior. Our approach uses a transformer-based reinforcement learning algorithm to train an agent that acts as a human observer, with the primary role of watching videos and simulating human gaze behavior. We employed an eye-tracking dataset gathered from videos generated by the VirtualHome simulator, with a primary focus on activity recognition. Our experimental results demonstrate the effectiveness of our gaze prediction method by highlighting its capability to replicate human gaze behavior and its applicability for downstream tasks where real human-gaze is used as input.
\end{abstract}



\maketitle

\section{Introduction}
\label{sec:intro}
Eye-tracking technology is increasingly being incorporated into various devices such as VR, AR, and smart glasses, due to recent advances. Its improved accessibility and precision allow for its wide use across different domains such as medical imaging~\cite{karargyris2021creation}, driver assistance systems~\cite{wu2019gaze}, and natural language processing~\cite{barrett2020sequence}. Gaze data is also employed in human-robot interaction~\cite{singh2020combining} and in action recognition tasks, mainly based on ego-centric videos~\cite{fathi2012learning,ogaki2012coupling,li2018eye}. These applications exploit valuable insights into the human visual engagement and perception that eye-tracking data offer.

Applications that utilize human gaze data for various tasks require automation to broaden their scope and to be functional in scenarios where human input is not available. The accurate prediction and emulation of human gaze behavior are crucial for facilitating these processes without human intervention. Current literature includes numerous studies on the prediction of gaze behavior. Nevertheless, these works primarily focus on ego-centric perspectives~\cite{li2013learning,huang2018predicting,zhang2017deep,rodin2021predicting}. In addition to these existing studies, it is essential to develop methods for predicting the human gaze in third-person view videos to gain insights into how individuals perceive and understand these videos. This information, similar to insights gained from ego-centric perspectives, is highly valuable. Yet, despite its significance, this area has been less explored in existing literature, emphasizing a considerable gap in research on third-person view gaze prediction.

Addressing this challenge, we have introduced a novel approach for predicting human gaze in third-person view videos. By leveraging a reinforcement learning strategy, we employed transformers to train an agent capable of observing videos and replicating human gaze behavior. This approach enables the integration of gaze predictions into existing applications, eliminating the need for real human input. To accomplish this, we utilized a reinforcement learning agent specifically trained to learn and adapt its policy for accurate and dynamic gaze prediction across the entire video. Recognizing our gaze sequence prediction task as a long sequence prediction problem, we employed transformers, leveraging their strong capability to handle extensive data sequences. Our model, trained with real human gaze sequences that observe an agent's actions in third-person view videos, processes video input and predicts the gaze sequence for the entire video. Our contributions can be summarized as follows:
\begin{itemize}
    \item We propose a novel reinforcement learning algorithm designed to predict human \textit{top-down} gaze in viewing videos, aimed at enhancing video understanding. 
    \item We benchmark our algorithm against other gaze prediction algorithms on the task of gaze prediction. The results show that our approach achieves high precision in predicting gaze.
    \item Employing action prediction as a case study, we evaluate the efficiency of our algorithm in extracting key information. The findings reveal that our method discovers crucial information more effectively than other models, thereby benefiting action prediction.
\end{itemize}

\section{Related Work}
\label{sec:related work}
\paragraph{Gaze Prediction}

Gaze prediction is essential in numerous application fields, such as action recognition and virtual reality environments. In the task of action recognition, gaze behavior prediction reflects human visual attention and simulates human perception. Its ability to identify action-related areas through human visual attention significantly enhances the performance of action recognition tasks. \cite{li2013learning} proposed a gaze prediction framework incorporating head and hand data from the egocentric camera wearer. \cite{fathi2012learning} utilized Support Vector Machines (SVMs) to predict gaze locations in the context of action recognition. In a different domain, \cite{palazzi2018predicting} introduced the Multi-branch algorithm, an innovative approach for driver gaze prediction tailored to both egocentric and car-centric views. This algorithm integrates RGB images, optical flow, and semantic segmentation, marking a significant advancement in the field of driver gaze prediction.

In virtual reality settings, gaze prediction is a prevalent topic, largely due to foveated rendering algorithms that offer high-quality rendering at the user's gaze point. These gaze prediction algorithms utilize a virtual egocentric view and are typically designed for applications with short time frames. For example, \cite{Zhang_2017_CVPR} introduced the Deep Future Gaze (DFG) for egocentric views, a GAN-based model employing spatial-temporal CNNs to predict future gaze positions in subsequent frames. \cite{SGaze_tvcg19} developed SGaze, a model that forecasts future gaze locations using eye-gaze data and eye-head coordination, achieving improved accuracy without the need for additional hardware. Expanding on this,  \cite{DGaze_gaze_prediction_tvcg20} presented DGaze, a CNN-based algorithm that enhances gaze prediction precision by incorporating dynamic object positions, head movements, and saliency features. Furthermore, \cite{fixationnet_bulling_tvcg21} proposed FixationNet, a neural network designed to predict near-future gaze locations in virtual environments, demonstrating enhanced performance in gaze prediction tasks.  \cite{gaze_prediction_dyn_str_manip_games_ieeevr16} focused on gaze prediction for task-oriented games, leveraging game variables and gaze locations to improve depth perception.

\paragraph{Reinforcement Learning on Gaze Prediction.}
The prediction of gaze locations can be approached as a sequential decision-making problem, commonly addressed with reinforcement learning (RL) algorithms. ~\cite{lv2020improving} introduced the concept of Reinforced Attention (RA) in the domain of driver gaze prediction. Their approach involved the integration of reinforcement learning techniques with the pre-existing fixation map prediction algorithm, Multi-branch~\cite{palazzi2018predicting}. The Reinforcement Attention model is trained to generate highly concentrated and accurate gaze maps, substantially improving the precision in predicting a driver's focal points. ~\cite{lv2020improving} utilized reinforcement learning as a regulatory mechanism, thereby enhancing the efficiency of gaze prediction in the complex and dynamic context of driving scenarios. Additionally, ~\cite{yang2020predicting} introduce an inverse reinforcement learning to predict conditioned scanpaths, employing dynamic contextual belief maps of object locations to represent individuals' internal belief states. In a similar context, ~\cite{baee2021medirl} propose to use maximum entropy deep inverse reinforcement learning to model visual attention exhibited by drivers in accident-prone situations.

\section{Methodology}
\label{sec:method}
This section presents the preliminary concepts and examines our gaze prediction method in detail.
\subsection{Preliminaries}
Our approach leverages the Transformer architecture, as trained through reinforcement learning methodologies. The Transformer, introduced by ~\cite{vaswani2017attention}, for sequence modeling tasks by emphasizing attention mechanisms and eliminating the need for recurrence and convolutions. It can capture long-range dependencies and contextual relationships within sequences, rendering it exceptionally suitable for complex sequence modeling tasks.

Reinforcement Learning (RL) is a type of machine learning. In RL an agent learns to make decisions by performing actions within an environment to maximize cumulative rewards~\cite{kaelbling1996reinforcement}. This learning process hinges on the agent's interactions with the environment, enabling it to identify the most rewarding actions through a trial-and-error approach. Transformer-based  reinforcement learning algorithms, as introduced by \cite{schmidhuber2019reinforcement} and \cite{chen2021decision}, have shown remarkable effectiveness in big sequence prediction. These algorithms employ a reward conditioning approach for sequence modeling tasks. This technique diverges from conventional RL methods by aligning the generation of future actions with the inferred desired reward outcomes, thereby effectively leveraging the Transformer architecture.

We utilized Decision Transformer \cite{chen2021decision}, which is based on conditional sequence modeling, as our RL agent. Traditional RL methods typically involve estimating value functions or calculating policy gradients. However, Decision Transformers determine the best actions by conditioning an autoregressive model on the expected reward, prior states, and actions, utilizing a Transformer architecture that enables autoregressive generation. As a result, this approach facilitates the generation of future actions that successfully achieve the intended outcome. Decision Transformers are built upon the GPT architecture \cite{radford2018improving}, which introduces modifications to the Transformer \cite{vaswani2017attention} structure to enable autoregressive generation

\subsection{Gaze Prediction}
\label{sec:method-rl}
With the goal of accurately simulating human gaze sequences and ensuring reliable generalization, we employ reinforcement learning (RL) techniques to train an agent. Individuals who watch a video, direct their eye movements towards areas more related to the given task, therefore RL agents can predict these gaze locations carry valuable information regarding their current video understanding tasks. 

We utilized the architecture proposed by ~\cite{chen2021decision}, incorporating several modifications. To enhance the processing of image frames, we employed a pretrained ResNet model~\cite{he2016deep}. This model is designed to extract feature representations from the frames, which are important for comprehending the visual content targeted by the human gaze. In our methodology, we maintained the original area targeted by the gaze with a specified crop size, $B$, and subsequently applied a black mask to the surrounding regions.

The RL agent, represented as $Q$, processes a sequence of 3-tuple inputs, denoted by $(\hat{R}_t, s_t, p_t)$. Within this structure, $ s_t $ corresponds to the current state of the environment, while  $p_t$ represents the action, defined as the gaze coordinates $p_t = (x_t, y_t)$. The term $ \hat{R}_t $ corresponds to the return-to-go at time $t$.  
For a given video sequence $\mathcal{I} = \{I_1, I_2, \dots, I_K\}$ with each frame $I_i \in \mathbb{R}^{W\times H \times 3}$, a visual encoder denoted by the function $\theta(\cdot)$ is applied to obtain state $s_t = \theta(I_{t}, \hat{p}_{t})$, where the input $(I_{t}, \hat{p}_{t})$ corresponds to the frame $I_{t}$ with a focus area identified by $\hat{p}_{t}$, which indicates the region of human gaze. This region is highlighted by masking the surrounding areas in the frame, thereby emphasizing the gaze-targeted content for analysis.

The state sequence is $\mathbf{S} = \{s_1, s_2, \dots, s_K\}$. The corresponding gaze sequence is denoted as $\mathbf{P} = \{(x_1, y_1), (x_2, y_2), \dots, (x_K, y_K)\}$ with $x_i$ indicates the coordinate of x-axis and $y_i$ of y-axis for each frame $I_t$. We processed the gaze coordinates  $(x_t, y_t)$ by normalizing them. The reward $r_t$ which can be described as the difference between the ground truth and the predicted gaze location can be computed as $r_t = -1 \times |(x_t, y_t) - (\hat{x}_t, \hat{y}_t) |$, where $(\hat{x}_t, \hat{y}_t)$ are the predicted gaze location, derived from $\hat{\mathbf{p}}_t$. The return-to-go, provided as a condition at time $t$, is defined as the sum of future rewards from time $t$ until the end of the trajectory, denoted as $\hat{R}_t= \sum_{t'=t}^{K} r_{t'}$. The input sequence for autoregressive training and generation can be represented as \begin{equation}
\tau_{\{1:K\}} = \{\tau_1,\tau_2, \dots, \tau_t, \dots, \tau_K\}
\end{equation}
where $\tau_t = (s_t, \mathbf{p}_t, \hat{R}_t)$. The output of the visual encoder from the video frame centered at the gaze position denoted as $s_t$, corresponds to states in the decision transformer. The gaze position $\mathbf{p}_t$ is utilized as actions. Subsequently, return-to-go $\hat{R}_t$ is calculated from rewards associated with the difference between predicted and actual human gaze positions. To predict the action at the next timestamp, $\hat{p}_{t+1}$, for a sequence of length $L$, the following equation is used:
\begin{equation}
\hat{\mathbf{p}}_{t+1} = Q\big( \tau_{\{t-L:t\}}\big)
\end{equation}
where the agent model is denoted by $Q$.

The objective is to maximize the cumulative return, denoted as $\mathbb{E} \big[\sum_{t=1}^{K} r_t\big]$. To achieve this goal, we minimize the mean squared error loss calculated by comparing the predicted action $\hat{\mathbf{p}}_{t}$ with the actual subsequent action $\mathbf{p}_{t}$ in the dataset. The loss is defined as
\begin{equation}
\label{eq:loss_mse}
\mathcal{L}_{MSE} = \frac{1}{N} \sum_{t=1}^{N} (\mathbf{p}_t - \hat{\mathbf{p}}_t)^2
\end{equation}


Within this framework, the states correspond to the video frames, while the actions represent the fixation locations associated with each state. Our reinforcement learning agent interacts with the environment by highlighting the gaze area through the application of masking. The ResNet model is employed as the visual encoder, and we have customized the action encoder in the architecture of the Decision Transformer to accept gaze locations, \( (x, y) \), as inputs. This setup implies that we are dealing with continuous actions within a bounded region, where the bounded region corresponds to the image size. Similarly, the final layer is customized for gaze prediction, configured to output a pair of values \( (x, y) \) representing the coordinates of the predicted gaze point.

\section{Experiments}
\label{sec:experiments}
We presented experiment results for gaze prediction to show the efficiency of our proposed model. Additionally, we apply our gaze prediction algorithm for downstream tasks where real-human gaze is used as input for instance action recognition and activity prediction. For this purpose, we have chosen to integrate and examine the action recognition model proposed in \cite{ozdel2024gazeguided} as a case study for our analysis.

\subsection{Dataset}
We evaluated our model's performance using a dataset~\cite{ozdel2024gazeguided} primarily collected for gaze-based action recognition and activity prediction tasks. This dataset, which includes videos from VirtualHome~\cite{puig2018virtualhome}, is specifically designed to enable virtual agents to conduct everyday household tasks. The reason behind selecting this particular dataset is twofold: firstly, it provides a third-person view, which diverges from the predominantly ego-centric videos found in existing literature. Secondly, each activity in this dataset is recorded using multiple static cameras, offering a variety of third-person perspectives. This approach not only increases the challenge for our model but also serves as an initial test of its generalizability. 

\begin{figure}[ht]
    \centering
    \includegraphics[width=0.45\textwidth]{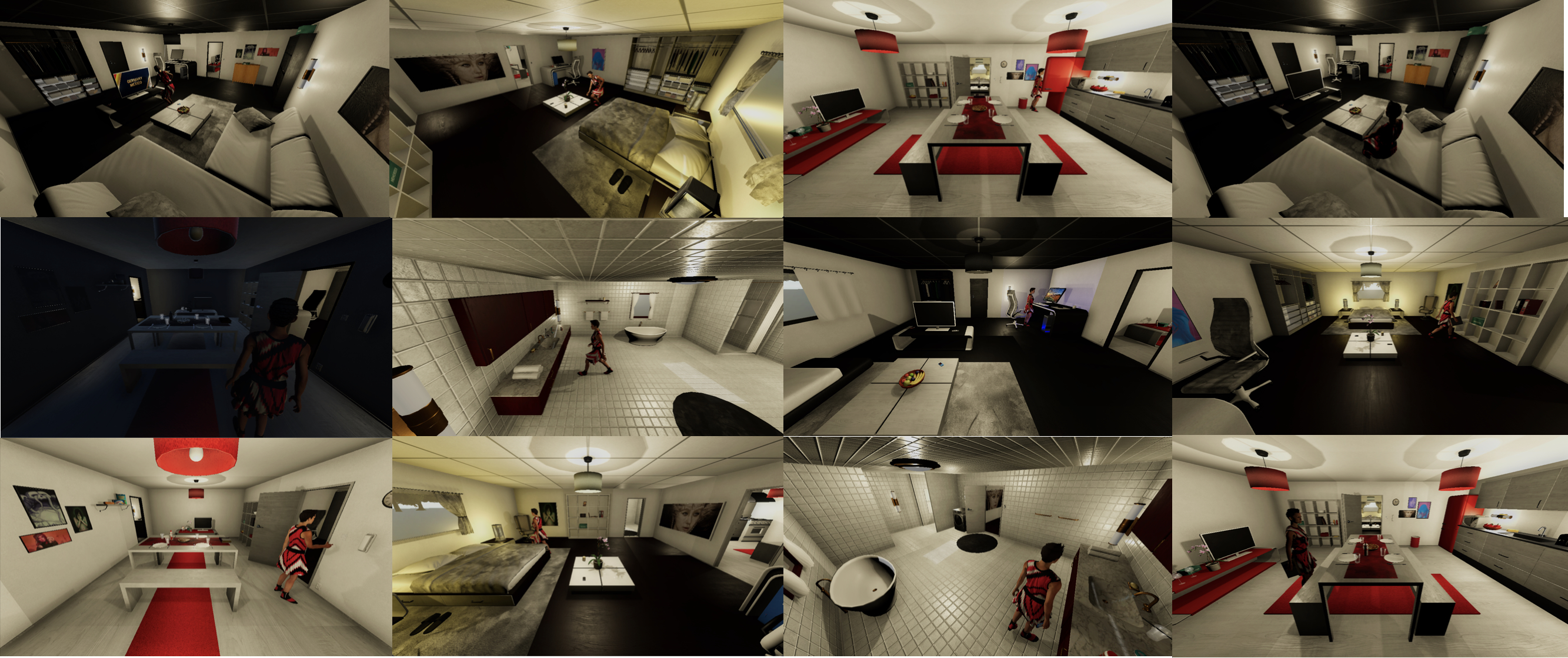}
    \caption{Sample frames from the eye-tracking dataset generated with the VirtualHome platform.}
    \Description{Sample frames from the eye-tracking dataset generated with the VirtualHome platform.}
    \label{fig:dataset-example}
\end{figure}

The dataset comprises 185 distinct videos for 18 different activities as programmed in~\cite{puig2018virtualhome}, with minor modifications. Each program includes a sequence of atomic actions, and every frame in the rendered videos is labeled with the corresponding atomic action. These activities take place within four different rooms: the living room, kitchen, bedroom, and bathroom. Figure~\ref{fig:dataset-example} presents some example frames from the video dataset generated using the VirtualHome platform.

To collect eye movement  data, we employed a Tobii Spectrum Eye Tracker, which operates at a sampling rate of 1200 Hz. The videos were displayed with a resolution of $1920 \times 1080$ pixels. Participants, seated approximately 60 cm from the screen, were asked to watch the VirtualHome videos and identify the activity class that best matched each video from the 18 available categories. This dataset encompasses eye movement data from 13 participants and comprises a total of 1311 videos with human gaze data. The dataset is partitioned into three subsets: 986 videos (approximately 75\%) for training, 60 videos (approximately 5\%) for validation, and 265 videos (approximately 20\%) for testing. Further details of this dataset can be found in ~\cite{ozdel2024gazeguided}.

\subsection{Implementation Details}

In our experiments, we adapted the Decision Transformer~\cite{chen2021decision} for gaze prediction based on minGPT \cite{minGPT}. Our model comprises a total of 6 transformer layers, each with an embedding size of 128 and 8 attention heads. We employed a context length of 128 for our agent and a highlighted area of size 75 pixels, corresponding to the human fovea area. We followed the guidelines in \cite{chen2021decision} and \cite{minGPT} for other hyper-parameters of our model. For image processing, the images were first resized to $224 \times 224$ pixels and then input into a pre-trained (on ImageNet \cite{deng2009imagenet}) ResNet-50 \cite{he2016deep}, which utilized as the visual encoder in our framework, followed by a linear layer to obtain state embeddings. The agent's training was conducted using the Adam optimizer with an initial learning rate of $1e^{-4}$, and it was trained for a maximum of 1000 epochs with a training batch size of 64. The source code will be publicly available on GitHub.

\subsection{Gaze prediction}
We show the advantages of our RL agent in gaze prediction compared to three baselines: (1) Random fixations sampled from normal distribution; (2) Behavioral cloning  (BC) and (3) Structured Gaze Modeling, a 3DCNN-based attention module from the framework Integrating Human Gaze into Attention~\cite{min2021integrating}. 
Structured Gaze Modeling deploys structured discrete latent variables to model gaze fixation points. Through the utilization of variational methods, this module learns to predict the distribution of gaze, which represents the state-of-the-art gaze prediction approach for long videos using CNNs. Additionally, it is specifically introduced for action recognition tasks in ego-centric views. Structured Gaze Modeling trains an attention module to minimize the cross-entropy loss using the Gumbel-Max reparameterization trick~\cite{maddison2014sampling} to predict discrete variables, in this case, gaze locations. In the Behavior Cloning (BC), we used the original model proposed by \cite{chen2021decision}. The exploration is facilitated by augmenting the actual human-gaze behavior data with Gaussian noise. Initially, this noise is introduced with high variance to facilitate exploration, then progressively reduced to lower values as the training advances. The model is trained using cross entropy loss in a supervised manner, in accordance with the approach described in~\cite{chen2021decision}.

To evaluate the gaze prediction performance, we use the Euclidean distance in pixels between the ground-truth and predicted gaze coordinates as well as the angular error between the ground-truth line of sight and the predicted line of sight following \cite{rolff2022gazetransformer}. Table \ref{tab:RL} shows the results, where our RL agent significantly outperforms all three compared methods. The improvement over BC shows that exploration in our RL is essential to train a successful gaze predictor. Compared to Structured Gaze Modeling, our RL predictor decreases the mean distance error by $37$ and $42$ pixels in the x-axis and y-axis, respectively, resulting in a notable difference in terms of perceived video content.
\begin{table}[t]
\caption{Results of fixation prediction. For both evaluation metrics, lower error indicates better performance. The best results are marked in bold.}
 \resizebox{0.85\linewidth}{!}{%
\begin{tabular}{c|c|c|c}
\toprule[1pt]
                
                & \multicolumn{2}{c|}{\textbf{Mean Distance Error} $\downarrow$} & {\multirow{2}{*}{\textbf{Mean Angular Error}$\downarrow$}}\\ \cline{2-3}
                &{X-axis}  &Y-axis &  \\ \hline
Random fixation & 350 & 332  & 12.43  \\\hline
BC  &    188     &   156    &   6.37       \\ \hline
Structured Gaze     &{154}        &    97    & 4.65          \\ \hline
\hline
Ours & \textbf{117}        &    \textbf{55}    & \textbf{3.31}         \\ 
\bottomrule[1pt]
\end{tabular}
}
\label{tab:RL}
\end{table}

\subsubsection{Qualitative results}
\begin{figure}[ht]
     \centering
     \begin{subfigure}{0.45\linewidth}
         \centering
         \includegraphics[width=\textwidth]{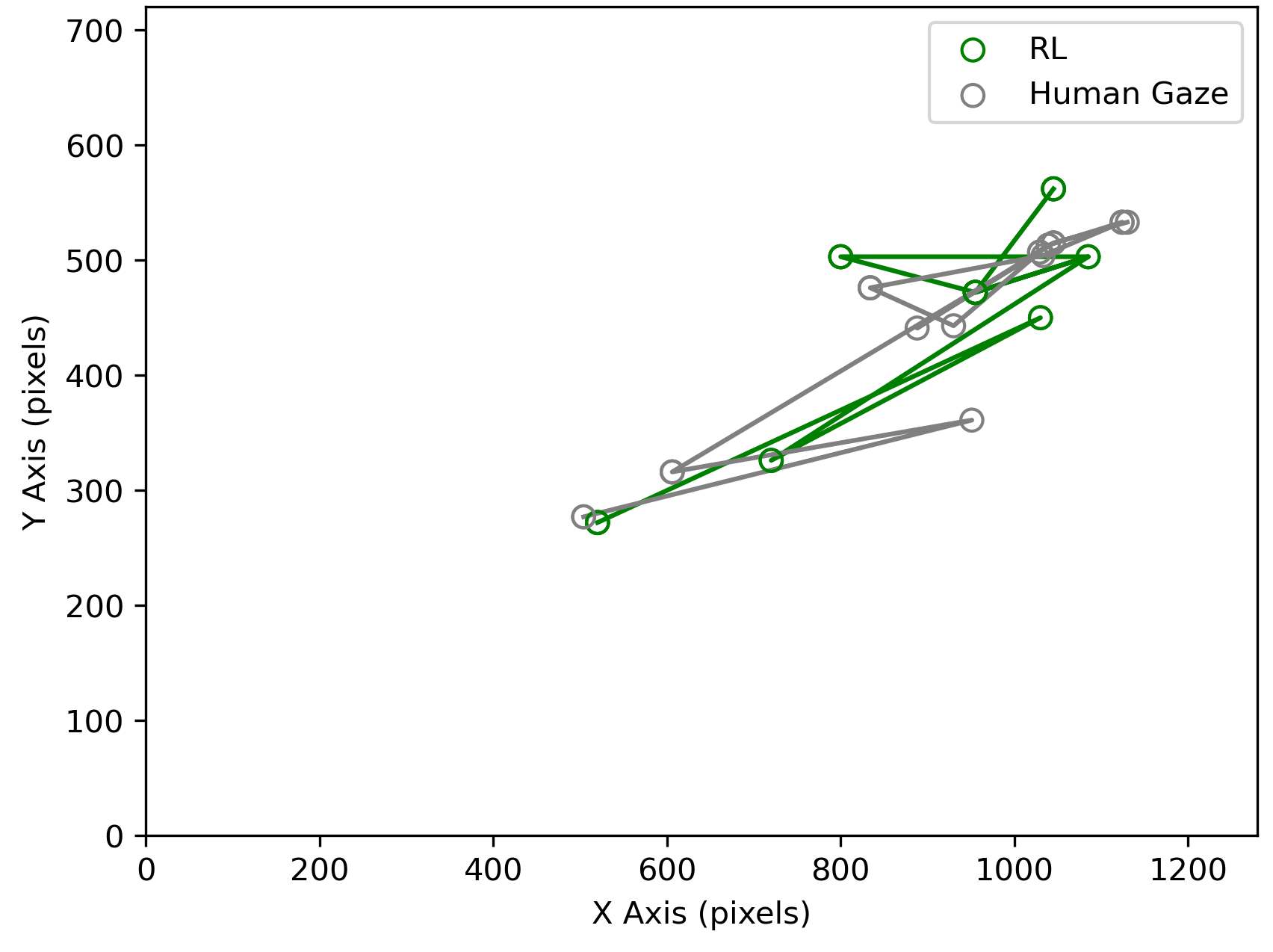}
     \end{subfigure}
     \hfill
     \begin{subfigure}{0.45\linewidth}
         \centering
         \includegraphics[width=\textwidth]{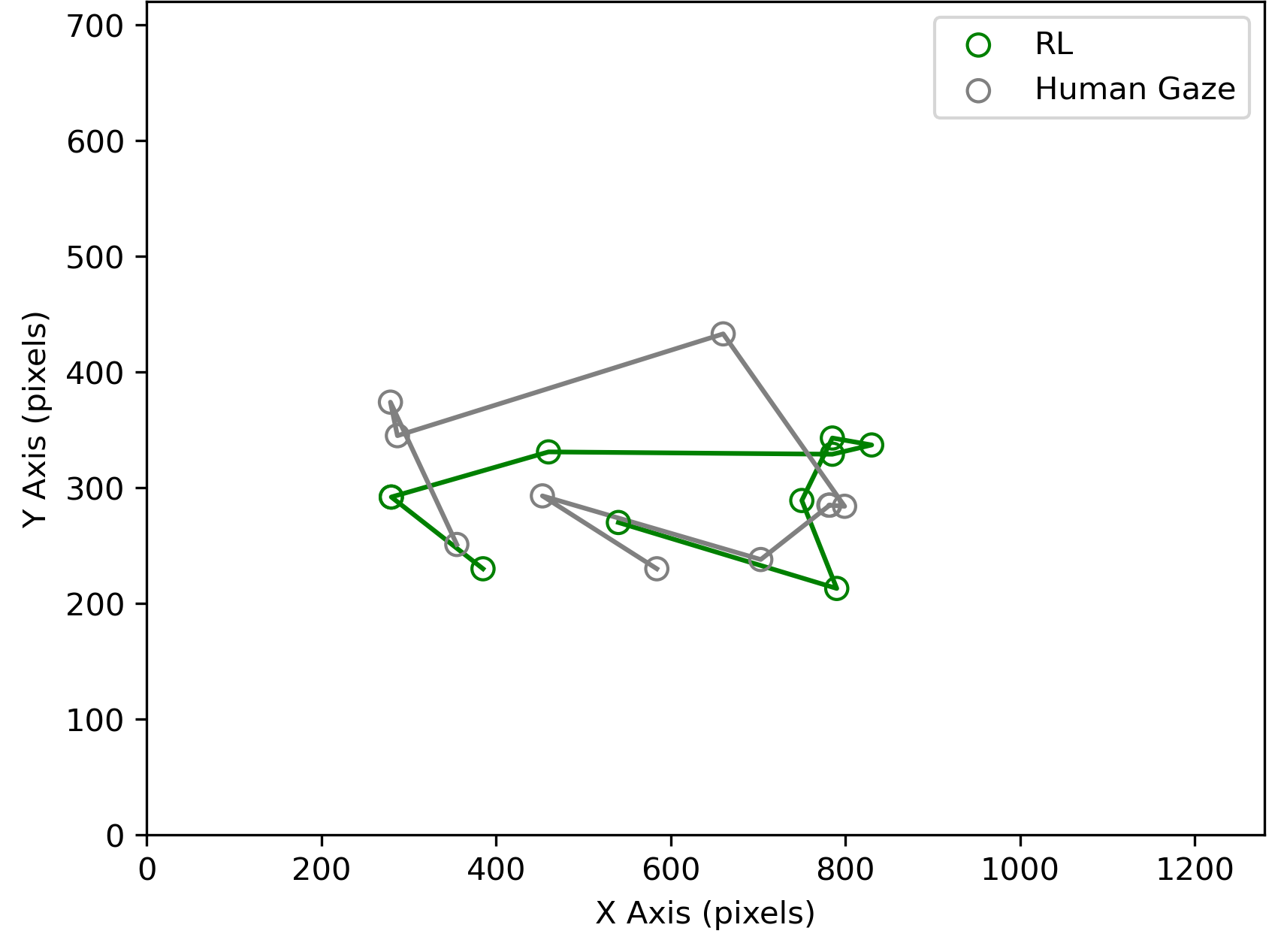}
     \end{subfigure}
        \caption{Trajectory comparison of ground-truth fixation (gray) and our prediction (green). Each dot represents a fixation point.}
        \Description{Trajectory comparison of ground-truth fixation (gray) and our prediction (green). Each dot represents a fixation point.}
        \label{fig:RL-quali}
\end{figure}

In Figure \ref{fig:RL-quali}, an example of RL prediction compared to the ground truth is demonstrated. For both, we show the trajectories by sampling ten frames to cover the fixation moves in a video from the test set. Our prediction (green line) follows the same trend as the ground-truth trajectory: they all explore the right upper part of the image as shown in the first example, while the left part of the image is explored in the second example. If we consider the foveal area, which is the perceived area with the fixation at the center, our predicted fixation discovers areas similar to those identified by the ground truth, an essential aspect in replicating human perception.

\subsection{Gaze prediction for action recognition task}
To demonstrate the effectiveness and practicality of our method, we have integrated it with an existing algorithm that is specifically designed for activity recognition and action prediction tasks, utilizing human-gaze input. We implemented the Gaze-guided Action Anticipation algorithm, as proposed in~\cite{ozdel2024gazeguided}. This approach constructs a visual-semantic graph from video inputs captured in a third-person view. It leverages a Graph Neural Network to identify the agent's intentions and forecast the sequence of actions required to achieve these intentions. By processing partial video sequences, the method comprehends the primary activity and anticipates future actions.

In this context, we replicated the settings outlined in the original paper of the methodology~\cite{ozdel2024gazeguided}. However, we utilized the gaze predicted by our RL agent instead of actual human gaze data. Thus, predicted gaze locations are utilized as input for the action recognition and action prediction model. We compared our integrated framework with four other state-of-the-art methods: I3D~\cite{carreira2017quo}, Ego-Topo~\cite{nagarajan2020ego}, IGA~\cite{min2021integrating}, and VideoGraph~\cite{hussein2019videograph}. The results for these methods, as presented in~\cite{ozdel2024gazeguided} for the corresponding dataset, were utilized. In addition, we included human performance data from the dataset for the activity classification task in Table \ref{tab:comparison}, which indicates a human accuracy of 94\%.

Each model analyzed 70\% of a video, followed by predicting the main goal and forecasting the action sequence required to accomplish this goal. To assess the effectiveness of integrating our approach with the existing methodology, we used classification accuracy as the metric for activity recognition. Atomic action prediction quality is evaluated using Intersection over Union (IoU) and normalized Levenshtein distance, accounting for sequence order. The IoU is defined as \(\text{IoU} = \frac{A \cap \hat{A}}{A \cup \hat{A}}\), comparing ground-truth (\(A\)) and predicted (\(\hat{A}\)) action sequences. Another crucial metric indicating the overall performance of these action prediction models is the task success rate on the VirtualHome platform. This measures the efficiency of the predicted action sequences. The success rate is calculated as the ratio of successfully executed tasks to the total number of test samples.

The Ego-Topo and Gaze-guided Action Anticipation models, which incorporate the human gaze, exhibit better performance over other approaches based on predicted gaze locations as shown in Table \ref{tab:comparison}. Our integrated framework, which utilizes predicted gaze locations, may not achieve the performance of human gaze-based methods but outperforms other techniques that rely on predicted gaze sequences. The original model~\cite{ozdel2024gazeguided}, utilizing human gaze fixations, achieves an activity recognition accuracy of $0.61$, and the final success rate of using the predicted atomic actions is $0.27$. When our RL is incorporated, it achieves an activity recognition accuracy of $0.40$ and a success rate of $0.18$ with the predicted action sequence. However, our integrated framework shows promising results, particularly for applications where human incorporation is not feasible, by achieving the best results among other comparable methods. Additionally, Ego-Topo, utilizing real human gaze input, achieves a success rate of $0.19$. Our integrated framework, which integrates an RL agent, attains a comparable success rate of $0.18$. This demonstrates that our proposed integrated approach can yield competitive outcomes in activity recognition and action prediction tasks, even without depending on the ground-truth human gaze data.

\begin{table}[ht]
\caption{Comparison of our RL agent integration to Gaze-guided Action Anticipation framework with other methods. $^*$ indicates that ground-truth gaze is used.}
\centering
\resizebox{.85\linewidth}{!}{
\begin{tabular}{c|c|c|c|c}
\toprule[1pt]
    & Acc. $\uparrow$ & IoU $\uparrow$ & Leven. $\downarrow$ &  SR $\uparrow$\\\hline
Human & 0.94 & - & - & - \\\hline
I3D & 0.12   &  0.03  & 0.79    & 0.06 \\ \hline
VideoGraph & 0.09  &    0.09  & 0.71   & 0.08 \\ \hline
IGA &  0.27  & 0.21 &   0.75 &  0.14     \\ \hline
Ego-Topo$^*$ &  0.54 &  0.26 & 0.63 &  0.19 \\
\midrule
\midrule
Gaze-guided Action Anticipation (HA)$^*$ & 0.61 & 0.35 & 0.51 & 0.27  \\ \hline
Gaze-guided Action Anticipation (RL) & 0.40 & 0.24 &0.69 & 0.18  \\ 
\bottomrule[1pt]
\end{tabular}
}
\label{tab:comparison}
\end{table}

As demonstrated in Table \ref{tab:RL}, our approach not only enhances the accuracy of gaze prediction but also yields promising results, proving its applicability to downstream tasks like activity recognition and action prediction as shown in Table \ref{tab:comparison}. These tasks were originally designed to utilize human gaze input, highlighting the applicability of our model in adapting to scenarios initially intended for human interaction.

\section{Conclusion}

In this paper, we introduce a novel method for predicting human \textit{top-down} gaze in third-person videos, utilizing a transformer-based model integrated with reinforcement learning (RL). The integration of a transformer-based architecture, renowned for its ability to recognize long-term complex patterns and predict future ones, significantly enhances our model's performance. Its combination with RL plays a substantial role in capturing and analyzing complex temporal dynamics. Our approach surpasses existing gaze prediction algorithms by offering more precise predictions that effectively mimic human gaze behavior throughout videos. The highly accurate gaze predictions generated by our model demonstrate its applicability in downstream tasks, such as action recognition and prediction. This indicates that our model is capable of effectively extracting key information, a crucial aspect in advancing video analysis techniques. Additionally, it reduces the need for human involvement in video-understanding tasks. In future work, we plan to explore the potential of our model in more complex and dynamic environments, aiming to further its applicability in real-world situations.

\begin{acks}
We are deeply grateful to IT-Stiftung Esslingen for their generous support of our hardware lab. 
\end{acks}

\bibliographystyle{ACM-Reference-Format}
\bibliography{ETRA_gaze}
\end{document}